\documentclass[letterpaper,10pt,conference]{ieeeconf}
\IEEEoverridecommandlockouts
\usepackage[T1]{fontenc}
\usepackage[utf8]{inputenc}
\usepackage{times}
\usepackage{amsmath,amssymb}
\usepackage{fix-cm}
\usepackage{graphicx}
\usepackage{booktabs,multirow,makecell,array}
\usepackage{cite}
\usepackage{url}
\usepackage{microtype}
\usepackage{balance}
\usepackage[table]{xcolor}
\usepackage{xspace}
\usepackage{algorithm}
\usepackage{algpseudocode}
\usepackage{tikz}
\usepackage{hyperref}
\usetikzlibrary{arrows.meta,positioning,calc}
\usepackage[caption=false,font=footnotesize]{subfig}
\usepackage[capitalize,noabbrev]{cleveref}
\newcommand{\tdmpcsq}{\mbox{TD-M(PC)\textsuperscript{2}}\xspace}
\newcommand{\tdmpcsqthree}{\mbox{TD-M(PC)\textsuperscript{2}\textsubscript{3M}}\xspace}
\newcommand{\Qrm}{\bar Q_{\bar\theta}^{\mathrm{rm}}}

\newcommand{\Lbase}{\mathcal L_{\pi}^{\mathrm{PC}}}

\newcommand{\ours}{\textsc{PL-MPC}\xspace}
\newcommand{\papertitle}{Beyond Policy Alignment: Closing the Planning–Learning Loop for Robot Control with Learned World Models}

\renewcommand{\arraystretch}{1.05}
\crefname{equation}{}{}
\Crefname{equation}{Equation}{Equations}
\crefname{section}{Sec.}{Secs.}
\Crefname{section}{Section}{Sections}
\crefname{figure}{Fig.}{Figs.}
\Crefname{figure}{Figure}{Figures}
\crefname{table}{Table}{Tables}
\Crefname{table}{Table}{Tables}
\newif\ifworkingappendix
\workingappendixfalse

\newcommand{\citep}[1]{\cite{#1}}
\usepackage[table]{xcolor}
\usepackage{bm}

\definecolor{bestcell}{RGB}{220,242,220}
\definecolor{secondcell}{RGB}{238,238,238}

\newcommand{\best}[1]{\cellcolor{bestcell}\(\bm{#1}\)}
\newcommand{\second}[1]{\cellcolor{secondcell}\(\underline{#1}\)}

\definecolor{bestcell}{RGB}{220,242,220}
\definecolor{secondcell}{RGB}{238,238,238}
\definecolor{poscell}{RGB}{230,245,230}
\definecolor{negcell}{RGB}{250,235,235}
\newcommand{\posdelta}[1]{\cellcolor{poscell}\(\bm{+#1}\)}
\newcommand{\negdelta}[1]{\cellcolor{negcell}\(-#1\)}
\definecolor{sectioncell}{RGB}{242,244,247}

\definecolor{componentcell}{RGB}{235,242,250}

\colorlet{oursrow}{bestcell}
\colorlet{refrow}{secondcell}
\newcommand{\mtd}{\texttt{MTD}}
\newcommand{\ate}{\texttt{ATE}}
\newcommand{\rad}{\texttt{RAD}}
\title{\LARGE\bf \papertitle}

\author{Kowndinya Boyalakuntla \qquad
        Yuhan Liu \qquad
        Abdeslam Boularias\\ [0.5em]
\small{Rutgers University}
}

\begin{document}
\maketitle
\thispagestyle{empty}
\pagestyle{empty}
\begin{abstract}
Planning with learned world models combines online trajectory optimization with learned value and policy functions for high-dimensional control. Because the planner determines the experience used for learning, while the learned critic and actor in turn score and propose future plans, planning and learning form a closed feedback loop. TD-MPC is a prominent instance of this design. Recent policy-constrained variants strengthen one part of the loop by aligning the learned policy with planner behavior. We introduce \ours (\emph{Planning--Learning MPC}), which additionally modifies critic supervision and planner terminal-value estimation. Hybrid multi-step TD targets expose critic updates to more realized rewards before bootstrapping; disagreement-aware terminal estimates reduce the influence of uncertain critic values during MPPI planning; and return-weighted actor distillation emphasizes planner-executed actions from high-return episodes. The world-model architecture and MPPI optimizer are otherwise unchanged. On HumanoidBench, the largest gains occur on \texttt{balance-hard}, where Total Average Return (TAR) increases from
$98\pm18$ to $387\pm255$, and \texttt{hurdle}, from $199\pm13$ to
$466\pm200$; performance across the broader benchmark remains task dependent,
and PL-MPC remains competitive on DMControl. Controlled ablations show
different component interactions across the two tasks.
We further
demonstrate zero-shot sim-to-real transfer on wrench--nut alignment with a
7-DoF KUKA IIWA14, obtaining higher observed success than \tdmpcsq on the
training object size and two unseen sizes. Code and data will be available at: \url{https://pl-mpc-humanoid.github.io}.
\end{abstract}
\section{Introduction}
\label{sec:intro}

Planning with learned world models has become a powerful approach to
high-dimensional control. A learned dynamics model predicts the consequences
of candidate actions, model-predictive control (MPC) optimizes short action
sequences online, and a learned value function estimates return beyond the
planning horizon. Many recent methods also learn a policy that supplies action
proposals and supports value learning. The TD-MPC family
~\cite{hansen2022tdmpc,hansen2024tdmpc2} is a prominent realization of this
architecture, combining latent world-model learning with Model Predictive Path
Integral (MPPI) control~\cite{williams2017mppi}.
This architecture creates a closed planning--learning loop. The online planner
generates experience; replayed experience updates the world model, critic, and
actor; the critic then returns to the planner as a terminal value estimator;
and the actor supplies both planning proposals and actions for TD
bootstrapping. Errors or mismatches at one interface can therefore influence
the data collected at the next. Recent TD-MPC variants have focused primarily
on the planner--policy interface by constraining or distilling the actor toward
planner behavior~\cite{lin2025tdmpcsq,wang2025bmpc,zhan2026bootstrap,
serra2026pompc}.

Policy alignment, however, addresses only one part of this loop. First,
standard one-step TD learning exposes the critic to only the immediate observed
reward before bootstrapping, so later rewards affect earlier values only
through subsequent updates. Second, the same learned critic supplies terminal
values for model-generated MPC rollouts, where uncertain estimates can change
trajectory ranking and hence future data collection. Third, planner-generated
experience can vary substantially in quality, yet planner-to-policy transfer
need not distinguish trajectories by their realized outcomes. These three
interfaces-critic supervision, terminal-value evaluation, and
planner-to-policy transfer motivate our approach.

\begin{figure}[t]
\centering
\includegraphics[width=1.0\linewidth]{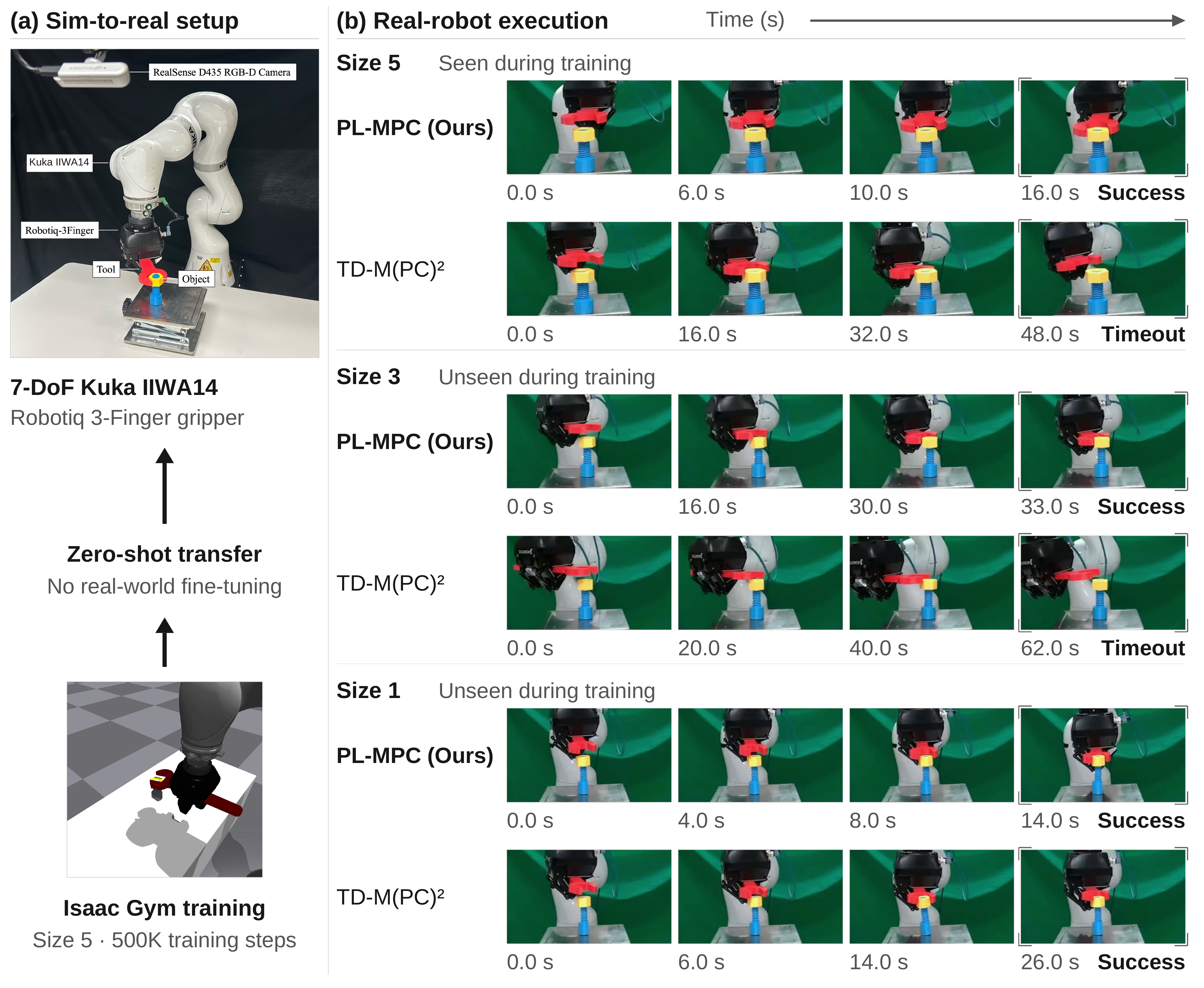}
\caption{\textbf{Zero-shot real-robot wrench--nut alignment.}
(a) PL-MPC is trained entirely in Isaac Gym and deployed on a KUKA IIWA14
without real-world fine-tuning; the deployed checkpoint is selected within
the first 500K training steps.
(b) Representative trials on the training size (size~5) and unseen
sizes~3 and~1. Frames are sampled at different times during each execution
and are not consecutive control steps. In the shown size~5 and size~3
trials, PL-MPC completes the task while \tdmpcsq reaches the 128-step limit.
On size~1, both methods succeed, with PL-MPC completing the trial in fewer
control steps. Successful sequences end with the wrench seated around the nut.}
\label{fig:real_setup}
\end{figure}

We introduce \ours (\emph{Planning--Learning MPC}), a simple extension of the
policy-constrained \tdmpcsq~\cite{lin2025tdmpcsq} backbone that modifies these three interfaces.
\emph{Multi-step TD targets} (\mtd{}) incorporate observed rewards from
planner-generated replay before the terminal bootstrap, using learned-model
predictions only when the target extends beyond the sampled replay slice. An
\emph{adaptive terminal estimate} (\ate{}) penalizes MPPI terminal values
when the target-critic ensemble disagrees, reducing reliance on uncertain value
estimates during planning. Finally, \emph{return-weighted actor distillation}
(\rad{}) weights imitation of planner-executed actions by realized
episode return rather than by the current critic. These changes leave the
latent world model, critic architecture, and MPPI optimization procedure
unchanged.

We evaluate \ours on thirteen HumanoidBench locomotion tasks
~\cite{sferrazza2024humanoidbench} and four high-dimensional DMControl tasks
~\cite{tassa2018deepmind}. The largest gains over \tdmpcsq occur on
\texttt{balance-hard}, where Total Average Return (TAR) increases from
$98\pm18$ to $387\pm255$, and \texttt{hurdle}, from
$199\pm13$ to $466\pm200$, while performance across the broader benchmark
remains task dependent. Leave-one-out ablations reveal different component
interactions across the two tasks, and a warmup study shows that \rad{} is
sensitive to its activation schedule.

We further evaluate zero-shot sim-to-real wrench--nut alignment on a 7-DoF
KUKA iiwa14. Policies trained entirely in Isaac Gym
~\cite{makoviychuk2021isaac} are deployed without real-world fine-tuning
using RGB-D-based 6D pose tracking~\cite{wen2024foundationpose}. PL-MPC
achieves higher observed success than \tdmpcsq on the training object size
and two unseen sizes.

Our contributions are:
\begin{itemize}
\setlength{\itemsep}{0pt}
\setlength{\parskip}{0pt}
\setlength{\parsep}{0pt}
\setlength{\topsep}{2pt}
    \item We formulate policy-constrained TD-MPC as a coupled
    planning--learning loop and introduce \ours, which modifies critic
    supervision, MPC terminal-value evaluation, and planner-to-policy
    transfer without changing the underlying world model or MPPI optimizer.

    \item Through controlled ablations and training diagnostics, we show that
    these mechanisms interact in a task-dependent manner, with the largest
    gains on the difficult HumanoidBench \texttt{balance-hard} and
    \texttt{hurdle} tasks.

    \item We demonstrate zero-shot sim-to-real wrench--nut alignment on a
    7-DoF KUKA IIWA14, including transfer to object sizes unseen during
    simulation training.
\end{itemize}
\section{Related Work}
\label{sec:related}

\subsection{Planning with learned models and value expansion.}
Learned world models support policy learning through imagined trajectories
~\cite{hafner2023dreamerv3} and online control through finite-horizon planning
with learned terminal values~\cite{sikchi2022loop,hansen2022tdmpc,
hansen2024tdmpc2}. POLO combines online trajectory optimization and value
learning while assuming access to an accurate internal dynamics model
~\cite{lowrey2018plan}. Palenicek et al.~\cite{palenicek2023valueexpansion}
find diminishing gains from longer value-expansion horizons even with oracle
dynamics; in their experiments, model-free value expansion performs comparably to model-based variants,
suggesting that model error is not the sole bottleneck. Our \texttt{MTD} update instead uses a hybrid target: replay
rewards from the planner-generated trajectory are used where available, and a
learned-model tail is introduced only when the target extends beyond the
sampled replay slice.

\subsection{Planner--policy alignment.}
Online MPC can generate behavior that differs from the actor used for policy
updates and value bootstrapping, creating a planner--policy mismatch
~\cite{sikchi2022loop}. Recent TD-MPC variants address this interface in
different ways. \tdmpcsq constrains the actor toward planner behavior stored
in replay~\cite{lin2025tdmpcsq}; BMPC imitates an MPC expert and refreshes
planner supervision through lazy reanalysis~\cite{wang2025bmpc}; BOOM combines
value maximization with value-weighted imitation of planner behavior
~\cite{zhan2026bootstrap}; and PO-MPC regularizes policy optimization toward an
adaptive planning prior~\cite{serra2026pompc}. These methods primarily improve alignment between planner behavior and the learned policy. \ours builds on \tdmpcsq but also changes critic
supervision and the terminal values used to rank MPPI trajectories.

\subsection{Outcome-weighted policy learning and uncertainty.}
Weighted policy-regression methods such as AWR, AWAC, CRR, and IQL emphasize
actions according to estimated value or advantage
~\cite{peng2019advantage,ashvin2020accelerating,wang2020critic,
kostrikov2021offline}. Guided policy search similarly transfers behavior from
a trajectory optimizer to a parameterized policy~\cite{levine2013guided}.
Most closely related to \texttt{RAD}, TDMPBC/SIRL weights self-imitation using
trajectory-level realized return for humanoid control~\cite{zhuang2025tdmpbc}.
\ours also uses realized episode return, but applies it specifically to
planner-executed replay actions within a policy-constrained TD-MPC loop.
Ensemble uncertainty has also been used to control value expansion. STEVE
weights value-expansion horizons according to model and value uncertainty
~\cite{buckman2018steve}. In contrast, \texttt{ATE} leaves the TD regression
target unchanged and instead penalizes the terminal critic value used to rank
MPPI trajectories. Thus, the three PL-MPC mechanisms act at distinct points
in the same planning--learning loop: critic supervision, terminal trajectory
evaluation, and planner-to-policy transfer.
\section{Preliminaries}
\label{sec:prelim}

\begin{figure*}
    \centering
    \includegraphics[width=0.9\linewidth]{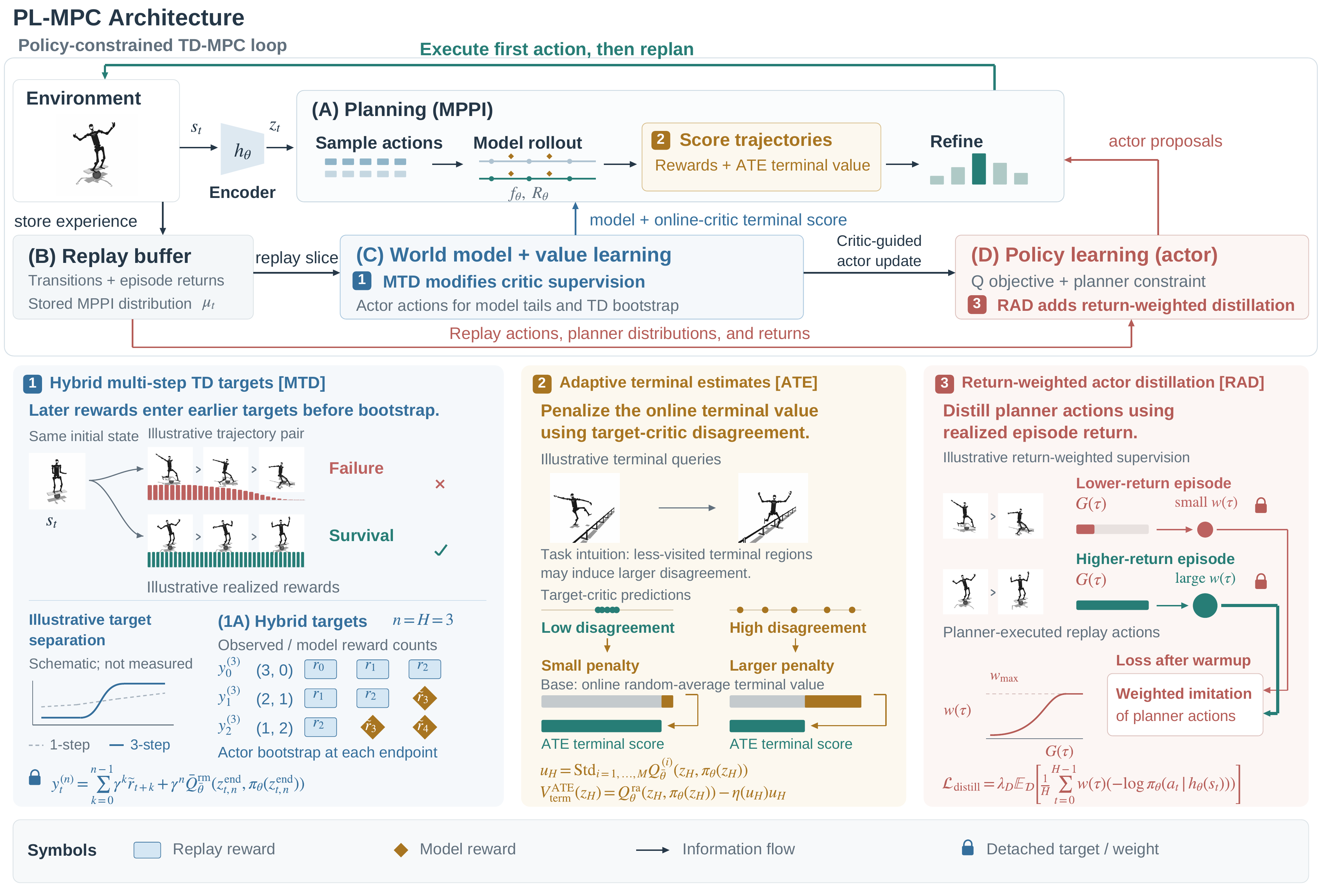}
\caption{\textbf{PL-MPC architecture.}
  The upper loop comprises MPPI planning (A), replay (B),
  world-model and value learning (C), and actor learning (D).
  Matching numbered badges link each modification to its detailed panel.
  \textbf{1: \mtd{}} constructs critic targets in C from replay
  rewards, using actor-driven model tails beyond the available replay
  slice; detail~1A shows the observed/model reward counts.
  \textbf{2: \ate{}} penalizes the planner's online random-average
  terminal value in A using target-critic ensemble disagreement.
  \textbf{3: \rad{}} uses realized episode returns to weight
  distillation of planner-executed replay actions in D.
  The actor supplies planning proposals and actions for model tails and
  TD bootstraps. Arrows show information flow; locks denote detached
  targets or weights.
  Dotted and solid target-separation curves illustrate one-step and
  multi-step TD supervision, respectively; these curves, ensemble
  spreads, and value/weight diagrams are schematic.}
    \label{fig:architecture}
\end{figure*}
\subsection{Control objective and latent world model}

We consider a discounted Markov decision process
$\mathcal{M}=(\mathcal{S},\mathcal{A},\mathcal{P},r,\gamma)$ with continuous
state and action spaces. A policy $\pi(a\mid s)$ maximizes the expected
discounted return
$J(\pi)=\mathbb{E}_{\pi}[\sum_{t=0}^{\infty}\gamma^t r(s_t,a_t)]$.
Its action-value function $Q^\pi(s,a)$ is the expected return after taking
$a$ in $s$ and following $\pi$ thereafter.
TD-MPC~\cite{hansen2022tdmpc,hansen2024tdmpc2} learns a control-oriented
latent world model. An encoder $h_\theta$ maps observations to latent states
$z_t=h_\theta(s_t)$, a dynamics model $f_\theta(z_t,a_t)$ predicts the next
latent state, and a reward model $R_\theta(z_t,a_t)$ predicts immediate
reward. These components are trained for control rather than observation
reconstruction. A stochastic actor $\pi_\theta(a\mid z)$ and an ensemble of
$M$ action-value functions $\{Q_\theta^{(i)}\}_{i=1}^{M}$ are learned
alongside the world model.

\subsection{Planning and temporal-difference learning}

At each environment step, TD-MPC performs online trajectory optimization
with Model Predictive Path Integral (MPPI) control~\cite{williams2017mppi}.
For a candidate sequence $a_{0:H-1}$, the learned model rolls out a latent
trajectory and scores it by
\[
\sum_{h=0}^{H-1}\gamma^h R_\theta(z_h,a_h)
+\gamma^H V_{\mathrm{term}}(z_H),
\qquad
z_{h+1}=f_\theta(z_h,a_h).
\]
The actor provides proposal trajectories, MPPI refines its sampling
distribution according to these scores, and only the first optimized action
is executed before replanning. The baseline planner terminal value is
$
V_{\mathrm{term}}(z_H)
=
Q_{\theta}^{\mathrm{ra}}(z_H,\pi_\theta(z_H)),
$
where the random-average operator $Q_{\theta}^{\mathrm{ra}}$ uniformly
samples two distinct heads from the online critic ensemble and averages
their decoded values. For compactness, $\pi_\theta(z)$ denotes an action
sampled from the stochastic actor when used as a critic input.
The critic is trained from replay. For a nonterminal transition, the
standard one-step TD target is
\begin{equation}
y_t^{(1)}
=
r_t
+
\gamma
Q_{\bar\theta}^{\mathrm{rm}}
\!\left(
z_{t+1},\pi_\theta(z_{t+1})
\right).
\label{eq:td0}
\end{equation}
Here $\bar\theta$ denotes target-critic parameters updated by Polyak
averaging of the online critic parameters. The random-min operator
$Q_{\bar\theta}^{\mathrm{rm}}$ uniformly samples two distinct target-critic
heads and returns the smaller of their decoded values. Thus, online critics
provide the planner terminal score, whereas target critics provide the TD
bootstrap.

Figure~\ref{fig:architecture} connects MPPI planning (A), replay (B),
world-model and value learning (C), and actor learning (D).
The replay action $a_t$ is selected by MPPI, whereas the continuation
action in Eq.~\eqref{eq:td0} is sampled from the actor.
Replay supplies training transitions through B$\rightarrow$C; the learned
world model and online critic then provide the rollout predictions and
terminal scores used by the planner in A.

\subsection{Policy-constrained TD-MPC}
\label{sec:backbone}

Our backbone is \tdmpcsq~\cite{lin2025tdmpcsq}, which constrains actor updates
toward the planner distribution stored in replay. Let
$\mu_t(\cdot\mid z_t)$ denote the MPPI proposal associated with a replay
transition. Suppressing implementation-specific normalization, the actor
objective is
\begin{equation}
\begin{aligned}
\Lbase
={}&
-\mathbb{E}_{a\sim\pi_\theta(\cdot\mid z)}
\left[Q_\theta(z,a)\right]
-\alpha\mathcal{H}\!\left[\pi_\theta(\cdot\mid z)\right]
\\
&+
\beta_{\mathrm{prior}}
\mathbb{E}_{a\sim\pi_\theta(\cdot\mid z)}
\left[-\log\mu_t(a\mid z)\right].
\end{aligned}
\label{eq:tdmpcsq-actor}
\end{equation}
The first term uses the online critic to favor high-value actions
(C$\rightarrow$D in Fig.~\ref{fig:architecture}), while the entropy
term regularizes the stochastic actor. The planner-prior term uses
$\mu_t$ from replay (B$\rightarrow$D) to discourage actions outside
the stored planner distribution.

\section{Method}
\label{sec:method}

As shown in Fig.~\ref{fig:architecture}, \ours modifies the critic
target in C (\mtd{}, panel~1), the MPPI terminal value in A
(\ate{}, panel~2), and the actor objective in D
(\rad{}, panel~3).
The latent world model, critic-ensemble architecture, and MPPI
optimization procedure are otherwise unchanged.

\subsection{Hybrid multi-step TD targets (\mtd{})}
\label{sec:method_mtd}

The one-step target in Eq.~\eqref{eq:td0} incorporates only the immediate
observed reward before bootstrapping. We instead use
\begin{equation}
y_t^{(n)}
=
\sum_{k=0}^{n-1}
\gamma^k\widetilde r_{t+k}
+
\gamma^n
\Qrm\!\left(
z^{\mathrm{end}}_{t,n},
\pi_\theta(z^{\mathrm{end}}_{t,n})
\right).
\label{eq:mtd}
\end{equation}
Whenever the corresponding transition lies within the sampled replay slice,
$\widetilde r_{t+k}=r_{t+k}$ is the observed reward from the planner-executed
trajectory. If the target extends beyond the available slice, the remaining
rewards and latent states are obtained by rolling the learned dynamics and
reward models forward under the current actor. Accordingly,
$z^{\mathrm{end}}_{t,n}$ is either an encoded replay state or the endpoint of
this model rollout.

The target is therefore hybrid rather than a fully observed $n$-step
return. For the default $n=H=3$, the targets at replay-slice positions
$t=0,1,2$ contain $(3,0)$, $(2,1)$, and $(1,2)$ observed and
model-predicted rewards, respectively
(Fig.~\ref{fig:architecture}, detail~1A).
The failure and survival trajectories in panel~1 illustrate how later
observed rewards enter earlier critic targets directly, rather than
only through successive one-step TD updates.

We substitute $y_t^{(n)}$ for $y_t^{(1)}$ in the standard TD-MPC categorical
value loss, retaining its temporal weighting $\rho^t$ with $\rho=0.5$. The
constructed target is stop-gradient: gradients update the critic prediction
but do not propagate through the actor, target critic, or world-model
computations used to construct the label. All other world-model and critic
losses are unchanged.
\mtd{} changes the information supplied to value learning, not merely the
discount applied to the terminal bootstrap. The target remains off-policy:
its observed prefix follows historical planner actions, while its
model-predicted tail and terminal bootstrap follow the current actor.

\begin{algorithm}[t]
\caption{\ours training.}
\label{alg:training}
\scriptsize
\algrenewcommand\algorithmicindent{0.9em}
\begin{algorithmic}[1]
\Require replay buffer $\mathcal D$, online parameters $\theta$,
target parameters $\bar\theta$, training counter $k$


\State Interact with the environment using MPPI, applying \ate{} to the
terminal score via Eq.~\eqref{eq:adaptive-terminal}; execute the first action
and store the transition and planner proposal

\State Upon episode completion, compute $G(\tau)$, associate it with
the episode's transitions.

\State Sample a contiguous replay slice
$(s_{0:H},a_{0:H-1},r_{0:H-1},G(\tau),\mu_{0:H-1})\sim\mathcal D$

\State Compute the multi-step targets $y_t^{(n)}$ using
Eq.~\eqref{eq:mtd}

\State Update the encoder, dynamics, reward model, and critics using the
standard TD-MPC losses with $y_t^{(n)}$

\State Compute the backbone actor objective $\Lbase$ (Eq.~\eqref{eq:tdmpcsq-actor}) using the stored planner
proposals $\mu_{0:H-1}$

\State Update the return-statistics queue from returns in the sampled minibatch

\If{$k\geq k_{\mathrm{warmup}}$ and return statistics are available}
    \State Compute $w(\tau)$ using Eq.~\eqref{eq:return-weight}
    \State Compute $\mathcal L_{\mathrm{distill}}$ using
    Eq.~\eqref{eq:distill}
    \State $\mathcal L_\pi
    \leftarrow
    \Lbase+\mathcal L_{\mathrm{distill}}$
\Else
    \State $\mathcal L_\pi\leftarrow\Lbase$
\EndIf

\State Update the actor using $\mathcal L_\pi$
\State Polyak-update target parameters $\bar\theta$

\end{algorithmic}
\end{algorithm}
\subsection{Adaptive terminal estimates (\ate{})}
\label{sec:method_ate}
\ate{} changes terminal evaluation within the MPPI planner in A
(Fig.~\ref{fig:architecture}, panel~2), using target-critic disagreement
at model-generated terminal latents to modulate the penalty applied to the
online-critic terminal score.

For each terminal query, we compute
$
u_H =
\operatorname{Std}_{i=1,\ldots,M}
Q_{\bar\theta}^{(i)}
\!\left(z_H,\pi_\theta(z_H)\right),
$
and use
\begin{equation}
\begin{aligned}
V_{\mathrm{term}}^{\mathrm{ATE}}(z_H)
&=
Q_{\theta}^{\mathrm{ra}}
\!\left(z_H,\pi_\theta(z_H)\right)
-\eta(u_H)u_H, \\
\eta(u_H)
&=
\eta_{\max}
\sigma\!\left(
\frac{u_H-\mu_u}{\sigma_u+\epsilon}
\right).
\end{aligned}
\label{eq:adaptive-terminal}
\end{equation}
Here $\sigma(\cdot)$ is the logistic sigmoid, and $\mu_u,\sigma_u$
are exponentially weighted estimates of the mean and standard deviation
of terminal-query disagreement. The normalization scales the penalty
relative to recently observed disagreement.

\ate{} modifies MPPI trajectory scoring only; it does not alter the TD target in
Eq.~\eqref{eq:mtd}. Ensemble disagreement is used as an uncertainty proxy,
not as a calibrated estimate of critic error.

\subsection{Return-weighted actor distillation (\rad{})}
\label{sec:method_distill}

\rad{} adds a loss to actor learning in D using planner-executed
actions and episode returns from replay
(B$\rightarrow$D in Fig.~\ref{fig:architecture}, panel~3).
The backbone constrains actor-sampled actions through the stored
planner distribution $\mu_t$; \rad{} additionally fits the
executed actions $a_t$, weighted by their source episode's realized
return.

Each replay transition is associated with the undiscounted episodic return
$G(\tau)=\sum_{t=0}^{T_\tau-1} r_t$ of its source trajectory. To normalize
returns, we maintain a finite FIFO queue of episode returns encountered in
sampled replay minibatches. Returns are deduplicated within each minibatch
before being appended, but may re-enter the queue in subsequent updates.
We compute $G_{\mathrm{med}}$ as the queue median and
$G_{\mathrm{std}}$ as its population standard deviation, and define
\begin{equation}
w(\tau)
=
\operatorname{clip}
\left[
\exp\!\left(
\frac{G(\tau)-G_{\mathrm{med}}}
{\max(G_{\mathrm{std}},\epsilon)}
\right),
0,w_{\max}
\right].
\label{eq:return-weight}
\end{equation}
Distinct finite returns are appended in ascending order within each
minibatch; for an even-sized queue, $G_{\mathrm{med}}$ is the lower
middle value.
Higher-return trajectories therefore receive greater weight relative to
recent replay experience, while clipping limits the influence of outliers.
The additional actor loss is
\begin{equation}
\mathcal L_{\mathrm{distill}}
=
\lambda_D
\mathbb E_{\mathcal D}
\left[
\frac{1}{H}
\sum_{t=0}^{H-1}
w(\tau)
\left(
-\log\pi_\theta(a_t\mid h_\theta(s_t))
\right)
\right].
\label{eq:distill}
\end{equation}
The return weight is treated as fixed during this update. Unlike
Q-weighted planner imitation~\cite{zhan2026bootstrap}, the weighting signal is
the realized trajectory return rather than the current critic estimate.

The actor is optimized with
$\mathcal L_\pi=\Lbase+\mathcal L_{\mathrm{distill}}$
once $k\geq k_{\mathrm{warmup}}$ and return
statistics are available; otherwise, it is optimized with $\Lbase$.
Algorithm~\ref{alg:training} summarizes one training iteration.
The only additional replay annotation required by \ours is the episodic
return $G(\tau)$; \tdmpcsq already stores the planner proposal used by its
policy constraint.

\section{Experiments}
\label{sec:eval}

\begin{figure*}[htbp]
\centering

\scriptsize
\setlength{\tabcolsep}{3.2pt}
\renewcommand{\arraystretch}{1.07}

\resizebox{0.76\textwidth}{!}{%
\begin{tabular}{@{}lccccccc@{}}
\toprule
\textbf{Task}
& \textbf{DreamerV3}
& \textbf{TD-MPC2}
& \textbf{BMPC}
& \textbf{BOOM}
& \textbf{\tdmpcsq}
& \cellcolor{oursrow}\textbf{\ours\ (Ours)}
& \textbf{Target} \\
\midrule

\rowcolor{sectioncell}
\multicolumn{8}{@{}l}{\textbf{DMControl}} \\

\texttt{dog-stand}
& $41 \pm 9$
& $523 \pm 421$
& $932 \pm 21$
& \best{982 \pm 7}
& $832 \pm 99$
& \second{976 \pm 14}
& -- \\

\texttt{dog-trot}
& $11 \pm 3$
& $394 \pm 184$
& \best{953 \pm 10}
& \second{923 \pm 8}
& $899 \pm 39$
& $885 \pm 78$
& -- \\

\texttt{humanoid-stand}
& $5 \pm 1$
& $637 \pm 71$
& \best{955 \pm 8}
& $920 \pm 15$
& $929 \pm 15$
& \second{954 \pm 2}
& -- \\

\texttt{humanoid-walk}
& $2 \pm 0$
& $643 \pm 96$
& \second{933 \pm 20}
& $921 \pm 10$
& $867 \pm 63$
& \best{947 \pm 6}
& -- \\

\midrule

\rowcolor{sectioncell}
\multicolumn{8}{@{}l}{\textbf{HumanoidBench Locomotion}} \\

\texttt{walk}
& $161 \pm 45$
& $891 \pm 43$
& $651 \pm 34$
& \best{928 \pm 16}
& \second{922 \pm 8}
& $911 \pm 7$
& $700$ \\

\texttt{stand}
& $220 \pm 74$
& $754 \pm 132$
& $801 \pm 13$
& $915 \pm 26$
& \second{927 \pm 48}
& \best{933 \pm 3}
& $800$ \\

\texttt{run}
& $56 \pm 11$
& $196 \pm 122$
& $358 \pm 152$
& $596 \pm 63$
& \best{852 \pm 8}
& \second{658 \pm 355}
& $700$ \\

\texttt{crawl}
& $504 \pm 90$
& $822 \pm 97$
& \best{922 \pm 17}
& $855 \pm 50$
& \second{899 \pm 64}
& $855 \pm 31$
& $700$ \\

\texttt{maze}
& $117 \pm 6$
& $196 \pm 36$
& \second{348 \pm 9}
& $341 \pm 2$
& $347 \pm 9$
& \best{353 \pm 3}
& $1200$ \\

\texttt{stair}
& $43 \pm 11$
& $66 \pm 14$
& $445 \pm 201$
& \best{462 \pm 75}
& $387 \pm 110$
& \second{461 \pm 12}
& $700$ \\

\texttt{slide}
& $19 \pm 6$
& $210 \pm 39$
& $498 \pm 18$
& $858 \pm 43$
& \second{902 \pm 24}
& \best{910 \pm 7}
& $700$ \\

\texttt{sit-simple}
& $268 \pm 26$
& $359 \pm 228$
& $695 \pm 105$
& $882 \pm 44$
& \best{932 \pm 18}
& \second{910 \pm 39}
& $750$ \\

\texttt{sit-hard}
& $152 \pm 20$
& \second{744 \pm 165}
& $576 \pm 44$
& $740 \pm 209$
& \best{821 \pm 64}
& $635 \pm 233$
& $750$ \\

\texttt{pole}
& $123 \pm 34$
& $156 \pm 25$
& $676 \pm 45$
& \best{879 \pm 33}
& \second{874 \pm 124}
& \best{879 \pm 49}
& $700$ \\

\texttt{balance-simple}
& $15 \pm 1$
& $120 \pm 24$
& $613 \pm 78$
& \best{784 \pm 101}
& $599 \pm 389$
& \second{765 \pm 176}
& $800$ \\

\texttt{hurdle}
& $13 \pm 3$
& $86 \pm 26$
& $165 \pm 64$
& \second{331 \pm 45}
& $199 \pm 13$
& \best{466 \pm 200}
& $700$ \\

\texttt{balance-hard}
& $17 \pm 4$
& $102 \pm 17$
& \second{104 \pm 23}
& $102 \pm 11$
& $98 \pm 18$
& \best{387 \pm 255}
& $800$ \\

\bottomrule
\end{tabular}%
}
\caption{\textbf{Benchmark performance.}
Total Average Return (TAR; mean$\pm$std across seeds).
Green and gray cells denote the highest and second-highest row means,
respectively; ties share the best highlight.
Target denotes the HumanoidBench reference-return threshold and is distinct
from the environment-defined success metric.}
\label{tab:main_return}

\end{figure*}

We evaluate \ours on thirteen HumanoidBench locomotion
tasks~\cite{sferrazza2024humanoidbench} and four high-dimensional
DMControl tasks~\cite{tassa2018deepmind}. We first evaluate whether the
proposed changes to value learning, MPC terminal-value estimation, and
policy distillation improve tasks on which the \tdmpcsq backbone remains
weak, without changing its world-model architecture or MPPI planner.
We then analyze the two tasks with the largest improvements to determine
how individual components affect performance and training dynamics.
Finally, we examine the sensitivity of return-weighted distillation to its
activation time. Zero-shot physical deployment is evaluated separately in
Sec.~\ref{sec:real}.

\subsection{Experimental setup}
\label{sec:protocol}
For HumanoidBench, DreamerV3 and TD-MPC2 results are computed from the
publicly released 2M-step evaluation logs, while BMPC, BOOM, \tdmpcsq,
and \ours are trained locally for 3M environment steps under the same task
protocol. For every method, each seed's reported score is the mean return
over its final five evaluations, and tables report mean$\pm$std across seeds.
Local runs are evaluated every 50K steps using ten episodes. Unless otherwise stated, \ours uses $n=H=3$, \ate{} uses
$\eta_{\max}=0.5$ with decay $0.99$, and \rad{} uses a 256-entry
return-statistics queue with $w_{\max}=10$, $\epsilon=10^{-3}$,
$\lambda_D=0.5$, and a 200K-step warmup. This configuration is fixed
across the main benchmark rather than tuned per task. Several HumanoidBench
configurations exhibit substantial across-seed variability; comparisons
below therefore refer to mean performance.

\subsection{Benchmark performance}
\label{sec:main_results}

\begin{figure}[htbp]
\centering
\scriptsize
\setlength{\tabcolsep}{2.8pt}
\renewcommand{\arraystretch}{1.08}

\resizebox{0.8\columnwidth}{!}{%
\begin{tabular}{@{}lcccc@{}}
\toprule
&
\multicolumn{2}{c}{\textbf{\texttt{balance-hard}}}
&
\multicolumn{2}{c}{\textbf{\texttt{hurdle}}} \\
\cmidrule(lr){2-3}
\cmidrule(lr){4-5}

Variant
& TAR
& Succ.
& TAR
& Succ. \\
\midrule

\rowcolor{refrow}
\tdmpcsqthree{}
& $98 \pm 18$
& $0.00$
& $199 \pm 13$
& $0.00$ \\

\midrule

w/o \mtd{}
& $186 \pm 51$
& $0.00$
& $530 \pm 275$
& $0.47$ \\

w/o \ate{}
& $206 \pm 63$
& $0.00$
& $296 \pm 21$
& $0.00$ \\

w/o \rad{}
& $272 \pm 221$
& $0.15$
& $231 \pm 72$
& $0.00$ \\

\rowcolor{oursrow}
\textbf{\ours\ (Ours)}
& \textbf{$387 \pm 255$}
& \textbf{$0.19$}
& $466 \pm 200$
& $0.19$ \\

\bottomrule
\end{tabular}%
}
\caption{\textbf{Leave-one-out component ablation.}
Total Average Return (TAR) and environment-defined success rate (Succ.),
averaged over three seeds. Each variant removes one component from full
PL-MPC; when enabled, \rad{} uses the default 200K warmup.}
\label{tab:leave_one_out}
\end{figure}
\begin{figure*}[t]
\centering
\subfloat[\texttt{balance-hard}.]{
\includegraphics[width=0.485\textwidth]
{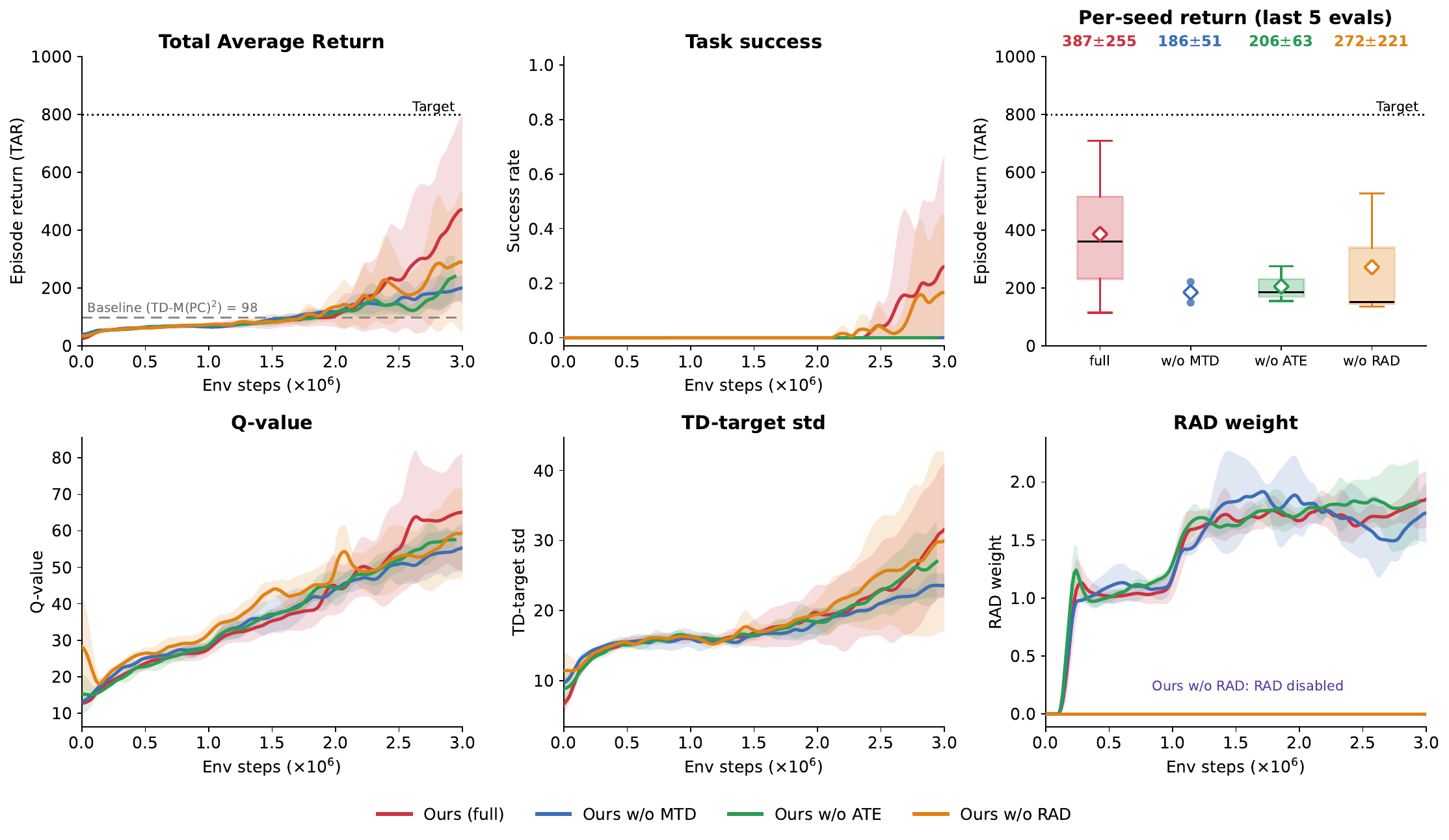}
\label{fig:loo_balance}}
\hfill
\subfloat[\texttt{hurdle}.]{
\includegraphics[width=0.485\textwidth]
{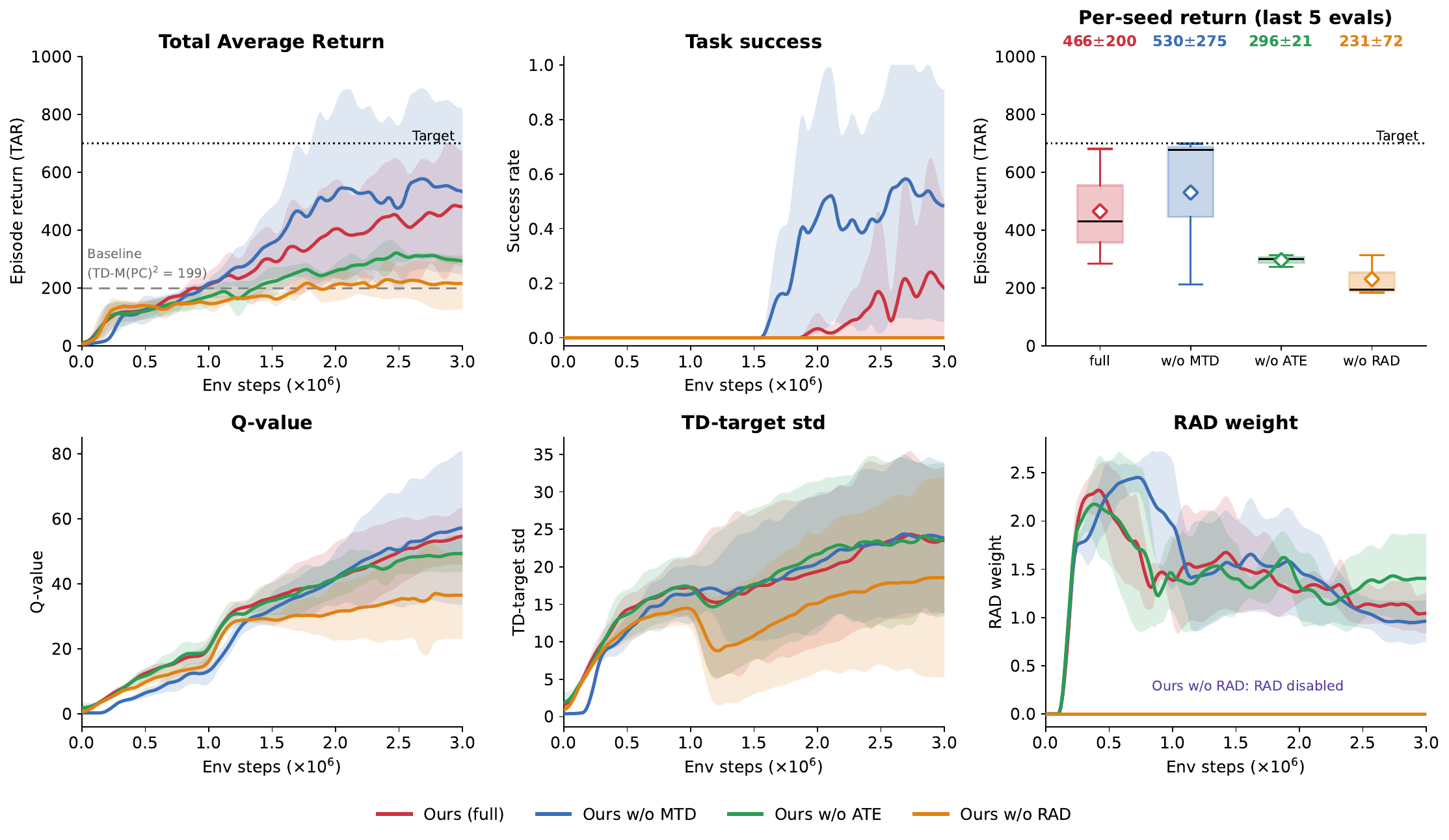}
\label{fig:loo_hurdle}}
\caption{\textbf{Leave-one-out ablations and training dynamics.}
Each panel compares full PL-MPC with variants that remove
\mtd{}, \ate{}, or \rad{}.
Top: TAR, environment-defined success, and per-seed return averaged over
the final five evaluations; dotted lines denote HumanoidBench
reference-return thresholds. Bottom: mean critic value, TD-target standard
deviation, and \rad{} weight. Curves show mean$\pm1$ standard deviation across
seeds. The TD-M(PC)$^2$ final return is shown for reference.}
\label{fig:leave-one-out}
\end{figure*}

PL-MPC shows its largest improvements over \tdmpcsq on
\texttt{balance-hard}, increasing TAR from $98\pm18$ to $387\pm255$, and
on \texttt{hurdle}, from $199\pm13$ to $466\pm200$
(Fig.~\ref{tab:main_return}). Both tasks exhibit substantial across-seed
variability, so these results reflect improvements in mean performance
rather than consistent gains across all seeds.


Across the thirteen HumanoidBench locomotion tasks, PL-MPC improves the
mean over \tdmpcsq on eight tasks and is lower on five. Of those five,
PL-MPC nevertheless exceeds the HumanoidBench reference-return threshold on
\texttt{walk}, \texttt{crawl}, and \texttt{sit-simple}; on these tasks, the
gap to \tdmpcsq is modest, averaging 26 TAR. The larger degradations are
concentrated on \texttt{run} and \texttt{sit-hard}, where PL-MPC also remains
below the reference threshold. Among all compared methods, PL-MPC attains the
highest mean TAR on \texttt{stand}, \texttt{maze}, \texttt{slide},
\texttt{hurdle}, and \texttt{balance-hard}, and ties BOOM on
\texttt{pole}. 

On the four
DMControl tasks with complete baseline coverage, PL-MPC remains competitive,
with the highest mean on \texttt{humanoid-walk}. We therefore focus the
analyses below on \texttt{balance-hard} and \texttt{hurdle}, where the
changes relative to the backbone are largest.

\subsection{Component ablations and training dynamics}
\label{sec:ablations}

The three modifications affect different parts of the TD-MPC update:
\mtd{} changes the critic target, \ate{} changes the terminal
value used for MPC trajectory scoring, and \rad{} adds
return-weighted policy distillation from planner-executed actions.
We evaluate their conditional contributions on \texttt{balance-hard} and
\texttt{hurdle} by removing one component at a time from full PL-MPC.
Fig.~\ref{tab:leave_one_out} summarizes final performance, while
Fig.~\ref{fig:leave-one-out} shows the corresponding learning curves and
training diagnostics.

On \texttt{balance-hard}, removing any component reduces mean TAR, with
the largest reduction occurring without \mtd{}
($387\pm255$ to $186\pm51$). As shown in
Fig.~\ref{fig:loo_balance}, the variants remain relatively close early in
training and diverge later, when some full-PL-MPC seeds reach substantially
higher returns. The accompanying critic and TD-target statistics characterize training
dynamics rather than value-estimation accuracy.

The component effects differ on \texttt{hurdle}. Removing \ate{} or
\rad{} reduces mean TAR to $296\pm21$ and $231\pm72$,
respectively, whereas removing \mtd{} yields $530\pm275$ and
$0.47$ success, exceeding the mean performance of full PL-MPC in this
experiment. The component effects are therefore task dependent and non-additive:
on \texttt{hurdle}, removing \mtd{} increases the observed mean, whereas
removing \ate{} or \rad{} lowers it. 

\subsection{Targeted ablations and training diagnostics}
\label{sec:diagnostics}

\begin{figure*}[t]
\centering
\subfloat[\texttt{balance-hard}: effect of \mtd{}.]{
\includegraphics[width=0.485\textwidth]{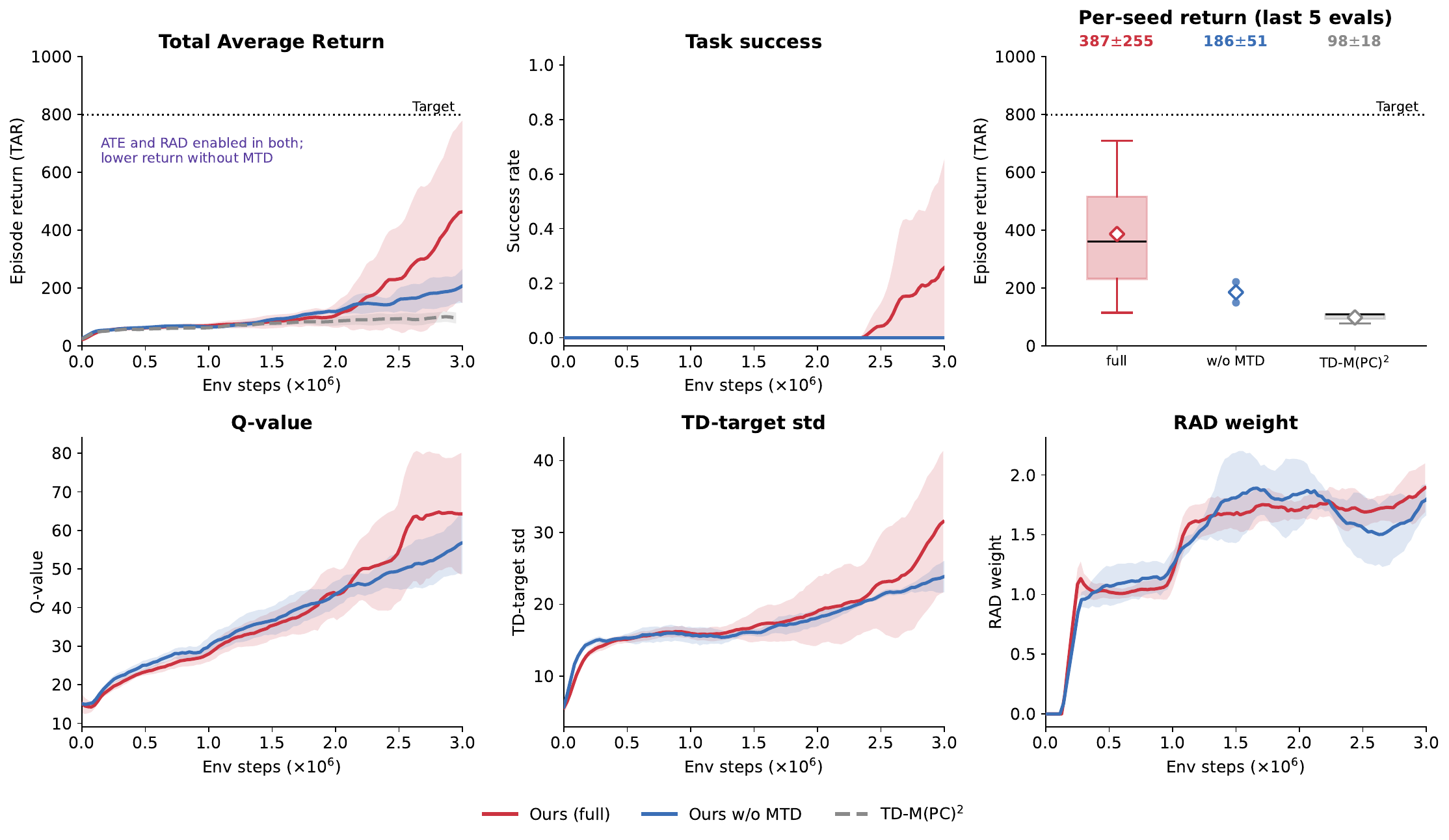}
\label{fig:mtd_isolation}}
\hfill
\subfloat[\texttt{hurdle}: effect of adding \ate{} and
\rad{} to \mtd{}.]{
\includegraphics[width=0.485\textwidth]{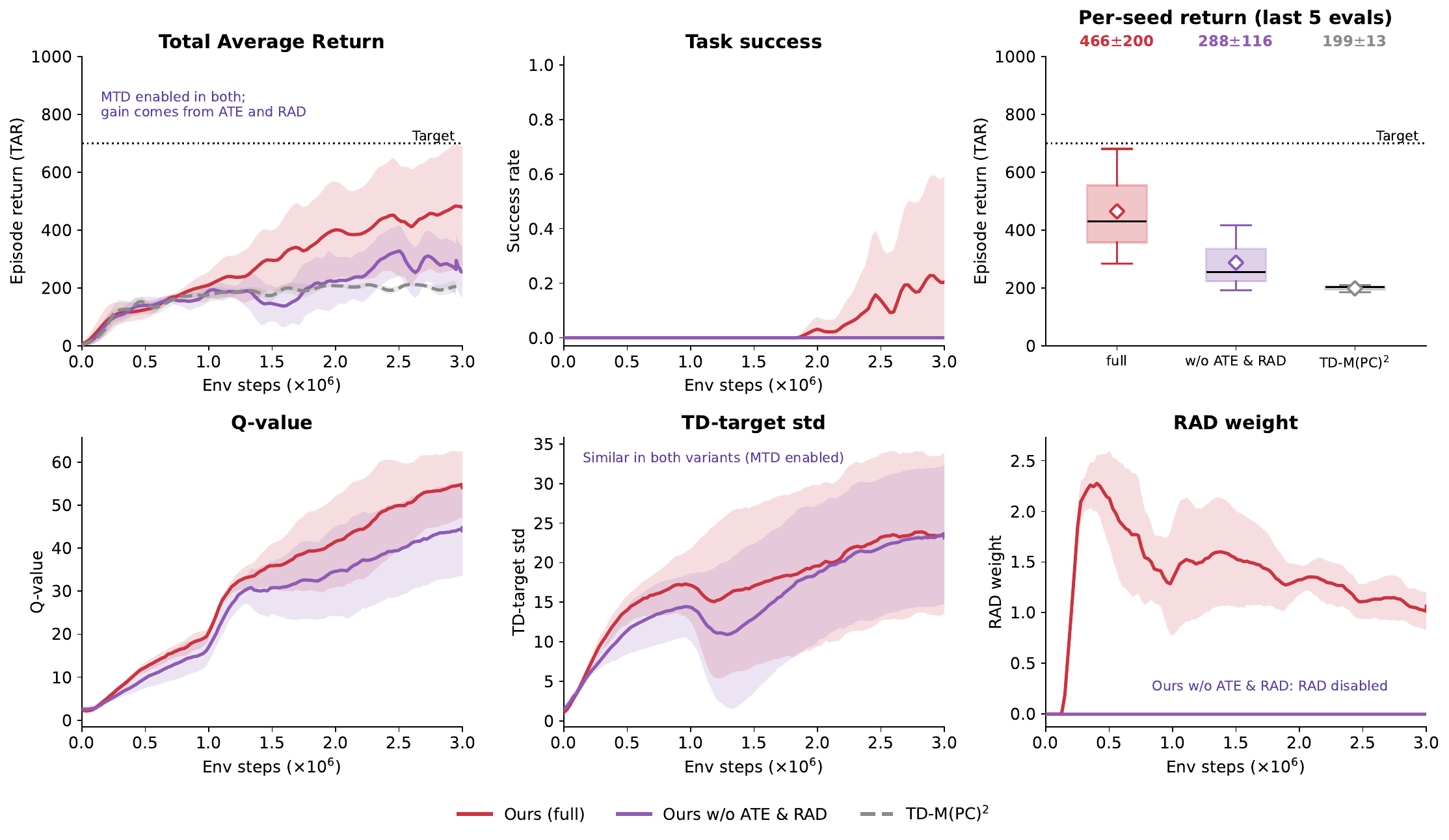}
\label{fig:hurdle_ate_rad}}
\caption{\textbf{Targeted ablations on the two analysis tasks.}
(a) On \texttt{balance-hard}, \ate{} and \rad{} are fixed and
only \mtd{} is removed.
(b) On \texttt{hurdle}, both variants use \mtd{}; the comparison adds
\ate{} and \rad{} jointly.
Top: episode return, task success, and per-seed return averaged over the
final five evaluations; dotted lines denote HumanoidBench reference-return
thresholds. Bottom: mean critic value, standard deviation of the TD target,
and \rad{} weight. The TD-M(PC)$^2$ backbone is shown in the performance
panels for reference. Diagnostic panels compare the corresponding PL-MPC
variants. Curves show mean$\pm1$ standard deviation across seeds.}
\label{fig:focused_ablation}
\end{figure*}

The leave-one-out study measures conditional component effects around full
PL-MPC. We next perform two targeted ablations that isolate specific component interactions and examine their training dynamics
(Fig.~\ref{fig:focused_ablation}).

\paragraph{\texttt{balance-hard}: effect of multi-step TD targets}
With \ate{} and \rad{} fixed, replacing the one-step TD target with the
hybrid $n=3$ target increases mean return from $186\pm51$ to $387\pm255$.
The difference is also reflected in task success: all seeds without
\mtd{} remain at $0\%$ success, whereas full PL-MPC averages $19\%$, with
the best seed succeeding on $50.9\%$ of evaluation episodes. On this task,
success requires maintaining balance for the full 1000-step episode, so the
successful seeds correspond to qualitatively different behavior rather than
a modest increase in accumulated return. The learning curves remain similar
early in training and separate later, when these successful behaviors emerge.
The critic-value and TD-target statistics characterize the accompanying
training dynamics rather than value-estimation accuracy.

\paragraph{\texttt{hurdle}: effect of \ate{} and \rad{} with \mtd{} fixed}
Both PL-MPC variants use \mtd{}. Adding \mtd{} alone to the \tdmpcsq
backbone yields $288\pm116$ mean return, compared with $199\pm13$ for
\tdmpcsq. Adding \ate{} and \rad{} jointly raises the mean to
$465\pm201$. The behavioral difference is also visible in task success:
the matched baseline seeds remain at $0\%$, whereas full PL-MPC averages
$19\%$ success and its best seed reaches $58.0\%$. Combined with the
leave-one-out ablation, these results show that the components interact:
\mtd{} improves over the backbone in this targeted comparison, whereas
removing \mtd{} from full PL-MPC does not reduce mean performance on
\texttt{hurdle}.









\subsection{Sensitivity to the \rad{} warmup}
\label{sec:warmup}

\begin{figure}[htbp]
\centering

\scriptsize
\setlength{\tabcolsep}{2.8pt}
\renewcommand{\arraystretch}{1.08}

\resizebox{\columnwidth}{!}{%
\begin{tabular}{@{}lccccc@{}}
\toprule

\textbf{Task}
& \cellcolor{refrow}\textbf{\tdmpcsq}
& \multicolumn{4}{c}{\cellcolor{oursrow}\textbf{\ours\ (Ours)}} \\
\cmidrule(lr){3-6}

&
& \textbf{200K}
& \textbf{750K}
& \textbf{$\Delta$}
& \makecell{\textbf{Succ.}\\\textbf{200K/750K}} \\
\midrule

\texttt{crawl}
& $899 \pm 64$
& $855 \pm 31$
& $971 \pm 6$
& \posdelta{116}
& $0.99/0.99$ \\

\texttt{maze}
& $347 \pm 9$
& $353 \pm 3$
& $349 \pm 8$
& \negdelta{4}
& $0.00/0.00$ \\

\texttt{walk}
& $922 \pm 8$
& $911 \pm 7$
& $910 \pm 1$
& \negdelta{1}
& $0.99/0.99$ \\

\texttt{balance-simple}
& $599 \pm 389$
& $765 \pm 176$
& $809 \pm 32$
& \posdelta{44}
& $0.63/0.65$ \\

\texttt{hurdle}
& $199 \pm 13$
& $466 \pm 200$
& $414 \pm 127$
& \negdelta{52}
& $0.19/0.09$ \\

\texttt{sit-hard}
& $821 \pm 64$
& $635 \pm 233$
& $805 \pm 104$
& \posdelta{170}
& $0.52/0.84$ \\

\bottomrule
\end{tabular}%
}
\caption{\textbf{Sensitivity to \rad{} warmup.}
Total Average Return (TAR; mean$\pm$std across three seeds) for the
\tdmpcsq backbone and PL-MPC with 200K and 750K \rad{} warmups.
$\Delta$ denotes the change from 200K to 750K; Succ.\ reports the
environment-defined success rates for the two warmups.}
\label{tab:warmup_success}
\end{figure}

PL-MPC with the default 200K \rad{} warmup underperforms its
\tdmpcsq backbone on five HumanoidBench tasks. Because 200K occurs early
in a 3M-step training run, distillation may begin before planner behavior has
sufficiently improved. We therefore test whether delaying \rad{} activation
changes these performance gaps. Fig.~\ref{tab:warmup_success} compares the
default 200K warmup with 750K on six tasks; the main benchmark retains the
fixed 200K setting.

A 750K warmup improves \texttt{crawl}, \texttt{balance-simple}, and
\texttt{sit-hard}, while \texttt{walk} and \texttt{maze} change little.
It decreases return and success on \texttt{hurdle}. Thus, no fixed warmup
dominates across these tasks. Delaying distillation closes the performance
gap to \tdmpcsq on \texttt{crawl} and substantially narrows it on
\texttt{sit-hard}. We retain 200K for the main benchmark and treat adaptive
scheduling of planner-to-policy distillation as future work.

\section{Real-Robot Wrench--Nut Alignment}
\label{sec:real}

\paragraph{Task and setup}
We evaluate zero-shot sim-to-real transfer on the wrench--nut alignment
benchmark introduced in~\cite{liu2025failure}. A 7-DoF KUKA IIWA14
equipped with a Robotiq 3-Finger gripper places a grasped wrench over a nut
threaded onto an upright bolt (Fig.~\ref{fig:real_setup}). Contact can rotate
the nut during execution, changing the desired wrench pose and requiring
closed-loop realignment. Policies are trained entirely in Isaac Gym~\cite{makoviychuk2021isaac} and
deployed without real-world fine-tuning. FoundationPose
~\cite{wen2024foundationpose} provides 6D object-pose estimates from a fixed
RealSense D435 RGB-D camera. Observations express the current and target
wrench poses in the nut frame, and actions specify an $\mathrm{SE}(3)$
displacement. Simulation randomizes the initial object configuration and
observation noise, while the low-level controller executes the commanded
end-effector displacements. We evaluate three matched nut--bolt--wrench asset sizes. Size~5, used during
simulation training, has a nut width across flats of $46$\,mm. Sizes~3 and~1
are unseen during training and measure $36$\,mm and $30$\,mm, respectively.
Nut height ($21.5$\,mm) and wrench thickness ($15.3$\,mm) are fixed across
the three sizes.

\paragraph{Simulation training}
Training episodes contain 256 control steps and start from the hardest
curriculum stage, with no curriculum progression during training.
Both methods
use the shaped reward
\[
r_t =
0.5\exp\!\left(
-\frac{d_t^2}{2(0.03)^2}
\right)
+0.3\,\mathbf{1}[\mathrm{engaged}]
+0.2\,\mathbf{1}[\mathrm{inserted}],
\]
where
\[
d_t =
\left\|
p_{\mathrm{head}}
-
\left(p_{\mathrm{nut}}+5\text{ mm}\,\hat z\right)
\right\|_2 .
\]
The \texttt{engaged} term requires
$|\Delta z|<10.75$\,mm and an $xy$ displacement below $50$\,mm, while
the \texttt{inserted} term requires $|\Delta z|<5$\,mm and an $xy$
displacement below $50$\,mm.
Simulation success uses a separate pose criterion. At the final step, the
wrench-head origin must lie within $\pm5$\,mm of the nut origin in height,
and the summed distance between four corresponding keypoints on the
wrench-head and nut axes must be below $50$\,mm. The criterion constrains
height, lateral alignment, and tilt while remaining invariant to rotation
about the nut axis.

For hardware deployment, we use one independently trained policy per method
and select its checkpoint using simulation evaluation within the first 500K
training steps. Policies are evaluated for ten MPC episodes every 10K steps,
and the checkpoint with the highest evaluation score is retained. For the
selected checkpoints, a separate 200-episode simulation evaluation on
training size~5 yields terminal-step success rates of $81.0\%$ for PL-MPC
and $67.5\%$ for \tdmpcsq; these episodes are not used for checkpoint
selection.

\paragraph{Hardware evaluation}
A trial is successful when the wrench head is inserted onto the nut before
the time limit; otherwise it terminates after 128 control steps. We conduct
31 trials per method on the training size and 15 trials per method on each
unseen size. For successful trials, we additionally report completion steps
and final position and rotation errors.
\begin{figure}[htbp]
\centering

\renewcommand{\arraystretch}{1.05}
\setlength{\tabcolsep}{2.5pt}

\resizebox{\columnwidth}{!}{%
\begin{tabular}{llcccccc}
\toprule
Method & Size & $N$ & SR (\%) & 95\% CI & Steps
& Pos. (mm) & Rot. (deg) \\
\midrule

\rowcolor{sectioncell}
\multicolumn{8}{l}{\textbf{Training object size}} \\

\tdmpcsq
& 5 & 31 & $61.3$ & $[43.8,\,76.3]$
& $53.3 \pm 24.4$
& $11.08 \pm 4.31$
& $7.28 \pm 4.26$ \\

\textbf{\ours~(Ours)}
& 5 & 31 & \best{74.2} & $[56.8,\,86.3]$
& $55.9 \pm 25.9$
& $9.86 \pm 4.50$
& $6.14 \pm 2.35$ \\

\midrule

\rowcolor{sectioncell}
\multicolumn{8}{l}{\textbf{Unseen object sizes}} \\

\tdmpcsq
& 3 & 15 & $73.3$ & $[48.0,\,89.1]$
& $54.1 \pm 16.0$
& $10.60 \pm 4.54$
& $3.57 \pm 1.92$ \\

\textbf{\ours~(Ours)}
& 3 & 15 & \best{80.0} & $[54.8,\,93.0]$
& $47.9 \pm 14.6$
& $11.57 \pm 4.41$
& $4.64 \pm 1.55$ \\

\addlinespace[2pt]

\tdmpcsq
& 1 & 15 & $20.0$ & $[7.0,\,45.2]$
& $52.0 \pm 7.2$
& $7.89 \pm 3.00$
& $3.69 \pm 1.86$ \\

\textbf{\ours~(Ours)}
& 1 & 15 & \best{33.3} & $[15.2,\,58.3]$
& $45.2 \pm 7.6$
& $9.75 \pm 3.78$
& $6.91 \pm 3.61$ \\

\cmidrule(lr){2-8}

\tdmpcsq
& Pooled & 30 & $46.7$ & $[30.2,\,63.9]$
& $53.6 \pm 14.3$
& $10.02 \pm 4.31$
& $3.60 \pm 1.84$ \\

\textbf{\ours~(Ours)}
& \textbf{Pooled} & 30 & \best{56.7} & $[39.2,\,72.6]$
& $47.1 \pm 12.8$
& $11.03 \pm 4.20$
& $5.31 \pm 2.46$ \\

\bottomrule
\end{tabular}%
}
\caption{\textbf{Zero-shot real-robot wrench--nut alignment.}
Success rate (SR) is computed over all trials with 95\% Wilson confidence
intervals. Steps and final position/rotation errors are
mean$\pm$sample standard deviation over successful trials.
Size~5 is used during simulation training; sizes~3 and~1 are unseen.
Pooled rows combine the two unseen sizes. \textbf{Bold green} denotes the
higher observed SR.}
\label{tab:real_results}
\end{figure}

\paragraph{Real-robot results}
PL-MPC achieves a higher observed success rate on all three object sizes:
$74.2\%$ versus $61.3\%$ on the training size, $80.0\%$ versus $73.3\%$
on unseen size~3, and $33.3\%$ versus $20.0\%$ on unseen size~1
(Fig.~\ref{tab:real_results}). Across the two unseen sizes, pooled success
increases from $46.7\%$ to $56.7\%$. PL-MPC also completes successful
trials in fewer control steps on both unseen sizes and in the pooled
comparison. Figure~\ref{fig:real_setup} shows representative executions,
including successful zero-shot transfer to both unseen sizes. Final pose
errors are comparable overall and do not favor either method consistently.
The hardware study demonstrates zero-shot sim-to-real deployment and
transfer to object sizes unseen during training, with the deployed PL-MPC
policy attaining a higher observed success rate than the deployed
\tdmpcsq policy on each tested size.
\section{Conclusions}
\label{sec:conclusion}

We presented \ours, which modifies critic supervision, MPC terminal-value
evaluation, and planner-to-policy transfer within policy-constrained TD-MPC. 
Its largest gains occur on HumanoidBench \texttt{balance-hard} and
\texttt{hurdle}, while improvements are less consistent on tasks such as
\texttt{maze}, where useful planner trajectories remain rare and
return-weighted distillation has little signal to amplify. HumanoidBench's
per-step survival reward can further obscure behavioral differences: policies
may accumulate substantial return by remaining stable without completing the
task, while successful seeds can occupy a distinct higher-return regime. This
behavior also contributes to the large across-seed variability observed on the
hardest tasks. On \texttt{balance-hard}, multi-step TD targets have the
largest leave-one-out effect, but outcomes remain strongly seed dependent. On
\texttt{hurdle}, the components interact differently; the targeted comparison
adds \ate{} and \rad{} jointly, while the leave-one-out study provides their
conditional effects rather than an additive decomposition.
Performance is also sensitive to when \rad{} is
activated: delaying distillation improves several tasks but degrades
\texttt{hurdle}, motivating adaptive rather than fixed warmup schedules. 
Zero-shot deployment further demonstrates transfer to the physical
wrench--nut task and to object sizes unseen during simulation training.

Although the proposed modifications are evaluated only on \tdmpcsq, they act
at interfaces shared by other TD-MPC-style methods, making systematic transfer
across planning--learning backbones a natural direction for future work. 

\addtolength{\textheight}{-12cm}
\bibliographystyle{IEEEtran}
\bibliography{references}

@inproceedings{hansen2022tdmpc,
  title     = {Temporal Difference Learning for Model Predictive Control},
  author    = {Hansen, Nicklas A. and Su, Hao and Wang, Xiaolong},
  booktitle = {Proceedings of the 39th International Conference on Machine Learning},
  series    = {Proceedings of Machine Learning Research},
  volume    = {162},
  pages     = {8387--8406},
  publisher = {PMLR},
  year      = {2022}
}

@inproceedings{hansen2024tdmpc2,
  title     = {{TD-MPC2}: Scalable, Robust World Models for Continuous Control},
  author    = {Hansen, Nicklas and Su, Hao and Wang, Xiaolong},
  booktitle = {International Conference on Learning Representations},
  year      = {2024}
}

@article{williams2017mppi,
  title   = {Model Predictive Path Integral Control: From Theory to Parallel Computation},
  author  = {Williams, Grady and Aldrich, Andrew and Theodorou, Evangelos A.},
  journal = {Journal of Guidance, Control, and Dynamics},
  volume  = {40},
  number  = {2},
  pages   = {344--357},
  year    = {2017},
  doi     = {10.2514/1.G001921}
}

@inproceedings{lin2025tdmpcsq,
  title     = {{TD-M(PC)$^2$}: Improving Temporal Difference {MPC} Through Policy Constraint},
  author    = {Lin, Haotian and Wang, Pengcheng and Schneider, Jeff and Shi, Guanya},
  booktitle = {Proceedings of the 8th Annual Learning for Dynamics and Control Conference},
  series    = {Proceedings of Machine Learning Research},
  volume    = {331},
  pages     = {705--736},
  publisher = {PMLR},
  year      = {2026}
}

@inproceedings{wang2025bmpc,
  title     = {Bootstrapped Model Predictive Control},
  author    = {Wang, Yuhang and Guo, Hanwei and Wang, Sizhe and Qian, Long and Lan, Xuguang},
  booktitle = {International Conference on Learning Representations},
  year      = {2025},
  url       = {https://openreview.net/forum?id=i7jAYFYDcM}
}

@inproceedings{zhan2026bootstrap,
  title     = {Bootstrap Off-policy with World Model},
  author    = {Zhan, Guojian and Wang, Likun and Zhang, Xiangteng and Gao, Jiaxin and Tomizuka, Masayoshi and Li, Shengbo Eben},
  booktitle = {Advances in Neural Information Processing Systems},
  volume    = {38},
  pages     = {148608--148636},
  year      = {2025}
}

@inproceedings{serra2026pompc,
  title     = {A {KL}-regularization Framework for Learning to Plan with Adaptive Priors},
  author    = {Serra-Gomez, Alvaro and Jarne Ornia, Daniel and Tirumala, Dhruva and Moerland, Thomas},
  booktitle = {Proceedings of the 43rd International Conference on Machine Learning},
  series    = {Proceedings of Machine Learning Research},
  volume    = {306},
  year      = {2026}
}

@inproceedings{sferrazza2024humanoidbench,
  title     = {{HumanoidBench}: Simulated Humanoid Benchmark for Whole-Body Locomotion and Manipulation},
  author    = {Sferrazza, Carmelo and Huang, Dun-Ming and Lin, Xingyu and Lee, Youngwoon and Abbeel, Pieter},
  booktitle = {Robotics: Science and Systems},
  year      = {2024},
  doi       = {10.15607/RSS.2024.XX.061}
}

@article{hafner2023dreamerv3,
  title   = {Mastering Diverse Control Tasks through World Models},
  author  = {Hafner, Danijar and Pasukonis, Jurgis and Ba, Jimmy and Lillicrap, Timothy},
  journal = {Nature},
  volume  = {640},
  pages   = {647--653},
  year    = {2025},
  doi     = {10.1038/s41586-025-08744-2}
}

@inproceedings{lowrey2018plan,
  title     = {Plan Online, Learn Offline: Efficient Learning and Exploration via Model-Based Control},
  author    = {Lowrey, Kendall and Rajeswaran, Aravind and Kakade, Sham and Todorov, Emanuel and Mordatch, Igor},
  booktitle = {International Conference on Learning Representations},
  year      = {2019}
}

@inproceedings{sikchi2022loop,
  title     = {Learning Off-Policy with Online Planning},
  author    = {Sikchi, Harshit and Zhou, Wenxuan and Held, David},
  booktitle = {Proceedings of the 5th Conference on Robot Learning},
  series    = {Proceedings of Machine Learning Research},
  volume    = {164},
  pages     = {1622--1633},
  publisher = {PMLR},
  year      = {2022}
}

@inproceedings{buckman2018steve,
  title     = {Sample-Efficient Reinforcement Learning with Stochastic Ensemble Value Expansion},
  author    = {Buckman, Jacob and Hafner, Danijar and Tucker, George and Brevdo, Eugene and Lee, Honglak},
  booktitle = {Advances in Neural Information Processing Systems},
  volume    = {31},
  year      = {2018}
}

@inproceedings{palenicek2023valueexpansion,
  title     = {Diminishing Return of Value Expansion Methods in Model-Based Reinforcement Learning},
  author    = {Palenicek, Daniel and Lutter, Michael and Carvalho, Joao and Peters, Jan},
  booktitle = {International Conference on Learning Representations},
  year      = {2023}
}

@article{peng2019advantage,
  title={Advantage-weighted regression: Simple and scalable off-policy reinforcement learning},
  author={Peng, Xue Bin and Kumar, Aviral and Zhang, Grace and Levine, Sergey},
  journal={arXiv preprint arXiv:1910.00177},
  year={2019}
}

@article{ashvin2020accelerating,
  title   = {{AWAC}: Accelerating Online Reinforcement Learning with Offline Datasets},
  author  = {Nair, Ashvin and Gupta, Abhishek and Dalal, Murtaza and Levine, Sergey},
  journal = {arXiv preprint arXiv:2006.09359},
  year    = {2020}
}

@inproceedings{wang2020critic,
  title     = {Critic Regularized Regression},
  author    = {Wang, Ziyu and Novikov, Alexander and Zolna, Konrad and Springenberg, Jost Tobias and Reed, Scott and Shahriari, Bobak and Siegel, Noah and Merel, Josh and Gulcehre, Caglar and Heess, Nicolas and de Freitas, Nando},
  booktitle = {Advances in Neural Information Processing Systems},
  volume    = {33},
  pages     = {7768--7778},
  year      = {2020}
}

@inproceedings{kostrikov2021offline,
  title     = {Offline Reinforcement Learning with Implicit Q-Learning},
  author    = {Kostrikov, Ilya and Nair, Ashvin and Levine, Sergey},
  booktitle = {International Conference on Learning Representations},
  year      = {2022}
}

@article{zhuang2025tdmpbc,
  title   = {{TDMPBC}: Self-Imitative Reinforcement Learning for Humanoid Robot Control},
  author  = {Zhuang, Zifeng and Shi, Diyuan and Suo, Runze and He, Xiao and Zhang, Hongyin and Wang, Ting and Lyu, Shangke and Wang, Donglin},
  journal = {arXiv preprint arXiv:2502.17322},
  year    = {2025}
}

@inproceedings{levine2013guided,
  title     = {Guided Policy Search},
  author    = {Levine, Sergey and Koltun, Vladlen},
  booktitle = {Proceedings of the 30th International Conference on Machine Learning},
  series    = {Proceedings of Machine Learning Research},
  volume    = {28},
  number    = {3},
  pages     = {1--9},
  publisher = {PMLR},
  year      = {2013}
}

@inproceedings{wen2024foundationpose,
  title     = {{FoundationPose}: Unified 6D Pose Estimation and Tracking of Novel Objects},
  author    = {Wen, Bowen and Yang, Wei and Kautz, Jan and Birchfield, Stan},
  booktitle = {Proceedings of the IEEE/CVF Conference on Computer Vision and Pattern Recognition},
  pages     = {17868--17879},
  year      = {2024},
  doi       = {10.1109/CVPR52733.2024.01692}
}

@misc{liu2025failure,
      title={Failure Forecasting Boosts Robustness of Sim2Real Rhythmic Insertion Policies}, 
      author={Yuhan Liu and Xinyu Zhang and Haonan Chang and Abdeslam Boularias},
      year={2025},
      eprint={2507.06519},
      archivePrefix={arXiv},
      primaryClass={cs.RO},
      url={https://arxiv.org/abs/2507.06519}, 
}

@inproceedings{
    makoviychuk2021isaac,
    title={Isaac Gym: High Performance {GPU} Based Physics Simulation For Robot Learning},
    author={Viktor Makoviychuk and Lukasz Wawrzyniak and Yunrong Guo and Michelle Lu and Kier Storey and Miles Macklin and David Hoeller and Nikita Rudin and Arthur Allshire and Ankur Handa and Gavriel State},
    booktitle={Thirty-fifth Conference on Neural Information Processing Systems Datasets and Benchmarks Track (Round 2)},
    year={2021},
    url={https://openreview.net/forum?id=fgFBtYgJQX_}
}

@article{tassa2018deepmind,
  title={Deepmind control suite},
  author={Tassa, Yuval and Doron, Yotam and Muldal, Alistair and Erez, Tom and Li, Yazhe and Casas, Diego de Las and Budden, David and Abdolmaleki, Abbas and Merel, Josh and Lefrancq, Andrew and others},
  journal={arXiv preprint arXiv:1801.00690},
  year={2018}
}
\end{document}